\documentclass[runningheads]{llncs}
\usepackage[T1]{fontenc}
\usepackage{float}
\usepackage{amsmath}
\usepackage{amsfonts}
\usepackage{subfig}
\usepackage{algorithm}
\usepackage{bbold}
\usepackage{algpseudocode}
\usepackage{hyperref}
\usepackage{multirow}
\usepackage{array}
\usepackage{graphicx}
\usepackage{xcolor}
\usepackage{adjustbox}
\usepackage{cite}
\definecolor{graph_cl_1}{RGB}{31, 119, 180}
\definecolor{graph_cl_2}{RGB}{255, 127, 14}
\definecolor{graph_cl_3}{RGB}{44, 160, 44}

\begin{document}
\title{Shrink the longest: improving latent space isotropy with simplicial geometry}
%
%
\author{Sergej Kudrjashov \orcidID{0000-0003-1899-4405} \\
 Olesya Karpik \orcidID{0000-0002-0477-1502} \\
 Eduard Klyshinsky \orcidID{0000-0002-4020-488\text{X}}}
 
%
%
\institute{National Research University “Higher School of Economics” \\
 Keldysh Institute of Applied Mathematics \\
 \email{xenomirant@gmail.com} 
}

\maketitle              
\begin{abstract}

Although transformer-based models have been dominating the field of deep learning, various studies of their embedding space have shown that they suffer from ``representation degeneration problem'' \textemdash \: embeddings tend to be distributed in a narrow cone, making the latent space highly anisotropic. Increasing the isotropy has been shown to improve performance in downstream tasks both in static and contextual language models. However, most approaches either add inference overhead or require a substantial amount of data for model reparametrization. We  propose a novel regularization technique based on simplicial geometry to improve the isotropy of latent representations. The core idea of our method is based on maximizing the persistent entropy of barcodes obtained using Vietoris-Rips filtration from contextual embeddings in the underlying latent space. We demonstrate that the method leads to an increase in downstream performance while significantly lowering the anisotropy during fine-tuning by exploiting existing geometric structures instead of reparametrization. The code is available at \url{github.com/Xenomirant/Shrink-the-longest}

\end{abstract}
\section{Introduction}

Since their introduction by Vaswani et al. \cite{Vaswani}, transformers have become the dominant SotA architecture in a variety of fields and applications. However, their impressive effectiveness and black-box nature made researchers wonder about their internal structure and seek reasons for their effectiveness and problems inherent to the class of models. Ethayarajh in \cite{Ethayarajh} notes that contextual word representations in ELMO, BERT and GPT-2 are highly anisotropic. Moreover, the extent of anisotropy grows with the number of layers \cite{Rajaee}. In other words, embeddings are not uniformly distributed in latent space, but clustered in a narrow cone, which limits the informational expressiveness of the space, making it impractical to be considered Euclidean anymore. As Gao et al. \cite{Gao} point out, it can be partly attributed to the optimization of rare tokens in models with tied embeddings \cite{Press}, which pushes them in the opposite direction to most of the hidden states. This is empirically justified by Liang et al. \cite{Yu_Guo_Liang}, who found a high correlation between embedding $L2$ norm and frequency in the pretraining corpora. However, it is far from explaining the whole story. Kawin Ethayarajh shows that less than 5\% of the variance in a word's contextual representations could be explained by the geometry of static embeddings \cite{Ethayarajh}.

Despite the continuing debate on sources of anisotropy, preserving isotropy has been shown to be a favorable property both theoretically and practically \cite{Lingxiao_Wang, Rajaee, Huang, Yu_Guo_Liang} -- not only in semantic similarity tasks, which seem to benefit the most from richer informational capacity of contextualized embeddings, but in terms of model convergence as well. It also helps to mitigate the "stolen probability effect" \cite{Demeter} -- a consequence of low embedding norm combined with overall positive cosine similarity between embeddings, which results in a failure to attribute high probability to certain token embeddings in any context.

Further investigation by Cai et al. \cite{Cai} has shown that while being globally anisotropic, the latent space consists of various clusters and manifolds that demonstrate significantly larger isotropy within themselves. This line of research was continued by Rajaee and Pilehvar \cite{Rajaee}, who found that most clusters are biased towards structural information, and e.g. verb tense representations are distributed in different subspaces. Thus, the clustering of the space is expressive and must be preserved, while we seek methods to improve overall isotropy.

Our contribution is the following. We address the problem by designing an additional regularization loss function inspired by recent results in computational topology \cite{Leygonie} that preserves the representational geometry of the latent space, while improving the isotropy by maximizing the entropy of distances between clusters. Thus we can optimize for isotropy while keeping the internal clustering structure intact, which differs from most previously proposed methods. Our method is, to our knowledge, the first one to leverage the differentiability of persistent barcodes to mitigate the anisotropy of the latent space. It should be noted that although we utilize it for pretrained language models, the method is in fact model-agnostic, and can be applied to any vector-based representational model.

The structure of the paper is as follows. In Section 2 we review the recent methods proposed for anisotropy mitigation. In Section 3 we describe the method and give a brief overview of results in computational topology we took inspiration from. Section 4 presents an experimental setup, while Section 5 is dedicated to the results we have obtained. Section 6 concludes the paper. 

\section{Related work}

To define and compute anisotropy we leverage the approach of Razzhigaev et al. \cite{Razzhigaev} and define it by exploiting a contextual embeddings' matrix decomposition.

Let $ X \in \mathbb{R}^{N \times D} $ represent the matrix of embeddings, where $\sigma_1, \dots, \sigma_{\min(N, D)}$ are its singular values, $N$ is the number of samples and $D$ is the hidden size. Then the $k$-th anisotropy score of $X$ is given by:

$$ anisotropy_k(X) = \frac{\sigma_k^2}{\sum_{i=1}^{\min(N, D)} \sigma_i^2} $$

Latest research has shown that anisotropy is a common trait in transformer-based architectures, where the overall contribution is split between attention \cite{Godey}, LayerNorm \cite{Gao, Xu} and various other factors.

There have been multiple methods proposed recently to address these peculiarities of transformers' latent space. A subsequent work of Gao et al. \cite{Gao} introduces explicit regularization of cosine similarity between normalized static embeddings. However, not only is its computation dependent on the vocabulary size, which appears to be costly for each gradient update, but it can also hardly be directly reapplied to subsequent layers, making it irrelevant to account for anisotropy induced by each of further layers. 

Another line of research was taken by Wang et al. \cite{Lingxiao_Wang}. They propose to reparametrize the initial embeddings matrix by a slow-decaying prior distribution over its singular values during training, resulting in a substantially flatter singular value spectrum. There have also been various attempts to adopt generative approaches for the task. Li et al. \cite{Li} utilized normalizing flows to map embeddings in the transformer's latent space to an isotropic Gaussian. Zhang et al. \cite{Zhang} explore the effect of using VAE with isotropic Gaussian posterior noting that it significantly improves the classification performance and robustness to input perturbations. 

However, despite their utility in developing higher isotropy, all these approaches seem to have a flaw that prevents them from being fully utilized during fine-tuning of already pretrained models. They reparametrize the whole internal space of the model, completely changing its geometry, which can result in suboptimal results when the amount of supervised data is small. A related finding is reported by Ding et al. \cite{Ding}. The authors evaluate the aforementioned isotropy calibration methods on various GLUE tasks finding that none of them produce consistent improvements over the uncalibrated models across tasks, domains and architectures.  They attribute it to the effectiveness of representations already found in local subspaces, which are utilized during fine-tuning.

\section{Method}

Topological data analysis has recently found lots of use cases in deep learning both due to its ability to represent global features, characteristic of the dataset, and due to recent results in differentiability of persistent homology \cite{Leygonie}, which allows one to construct topology-aware representations \cite{Trofimov, Moor}. We apply these results to construct our own regularization loss based on barcodes obtained from Vietoris-Rips complex of the embeddings during training.

\hspace{0.5cm}

\textbf{Definition}. Suppose $X$ is a metric space and $r > 0$ is a real number. The Vietoris-Rips complex at scale $r$ is defined as having X as the set of vertices (0-dimensional simplices) and a $k$-simplex for each set of $k+1$ vertices having diameter at most $r$, thus making it equivalent to a clique complex.

\hspace{0.5cm}

By varying $r$ we obtain a filtration over the simplicial complex, which provides us with additional information about the topological structure of the metric space. This gives us a powerful tool to explore the structure of high dimensional data without performing dimensionality reduction.

\hspace{0.5cm}

\textbf{Definition}. The persistent barcode of dimension $k$ is a multiset of intervals that mark ``births'' $x_i$ and ``deaths'' $y_i$ of $k$-dimensional holes detected by the $k$-th homology group as we vary the scale $r$.

\hspace{0.5cm}

As we are primarily interested in the local clustering structure of the space, we only take 0-dimensional homology into account, reducing the simplex to the construction of the minimum spanning tree, where individual bars represent the edge lengths of the tree. The number of bars is obviously dependent on the number of edges and equals $N - 1$, where $N$ is the number of vertices.

However, in general, not all the bars encode meaningful topological features, which motivates researchers to seek feature selection methods. We advance from the recent algorithm proposed by Atienza et al. \cite{Atienza} to define a functional that will serve us a dual role: as an optimization objective and as a tool for feature selection.

\hspace{0.5cm}

\textbf{Definition}. Given a filtration $F = \{K(t) | \: t \in \mathbb{R}\}$  and the corresponding diagram $dgm(F) = \{a_i = (x_i, y_i) | \: 1 \leq i \leq n\}$, let $L = \{l_i = y_i - x_i |\:  1 \leq i \leq n\}$. The persistent entropy $E(F)$ of $F$ is defined as the entropy of the distribution of bar lengths:

\begin{center}
$ E(F) = - \sum_{i=1}^n p_i \log(p_i)$,
\vspace{5px}
where $p_i$ is the normalized bar length, $S_L = \sum_{i=1}^n l_i, \: l_i = y_i - x_i, \: p_i = \frac{l_i}{S_L}$
\end{center}

The maximum persistent entropy corresponds to the situation when all the bars are of equal length. On the contrary, the value decreases as the distribution becomes more diverse. Note that this function is dependent on the number of bars and lies in the interval $[0, \log(n)]$.

This lets us set a base to describe the algorithm for prominent topological feature selection. For the empirical distribution of bars obtained from a batch let us suppose that the longest one $T$ and the tiniest one $r$ are the extreme cases of the observed distribution. $T$ is always considered a feature, while $r$ is considered noise. This assumption is crucial for the algorithm. 

\begin{algorithm}[h]
\caption{Separating topological features from noise (Atienza et al., 2019)}
\label{alg:cap}
\begin{algorithmic}
\Require A persistent barcode $B(VR_V) = \{(x_i, y_i) | 1 \leq i \leq n\}$ with bars of finite length
\vspace{3pt}
    \State $L = \{l_i:= y_i - x_i\}$
    \State $r \gets \min(L), \: T \gets \max(L), \: \alpha = \frac{r}{T}$
    \State $L \gets sort(L) = \{l_1, \dots, l_{n}\}$ s.t. $(l_n = T \geq l_1)$ \& $(l_i \geq l_j \geq l_{n-1} = r, \: 1 \leq i < j < n - 1)$ \Comment{Sort in decreasing order except for the last feature}
    \State $L_0^{\prime} \gets L, \: n^{\prime} \gets n$
    \Procedure{Select features}{$L_0^{\prime}$, $n^{\prime}$}:
        \For{$i=1 \: to \: i = n^{\prime}-2$}
        \State $P_i^{\prime} = \sum_{j=i+1}^n l_j, \: R_i^{\prime} = \{l_{i+1}, \dots, l_{n^{\prime}}\}, \: l_j^{\prime} = \frac{P_i}{e^{E(R_i)}} $
        \State $L_i^{\prime} = \{l_1^{\prime}, \dots, l_i^{\prime}, l_{i+1}, \dots, l_{n^{\prime}}\}$
    
        \State $C = \frac{S_{L_{i-1}^{\prime}}}{S_{L_i^{\prime}}} = \frac{P_{i-1}^{\prime} + (i-1) \times l_{i-1}^{\prime}}{P_i^{\prime} + i \times l_i^{\prime}}$
        \State $Q = \left \lceil{\frac{\alpha n^{\prime}(\alpha - 1 - \log(\alpha))}{(\alpha - 1)^2}}\right \rfloor $
        \If{$C \geq 1$}
            \State \textbf{Break}
        \EndIf
        \If{$Q \leq i$}
            \State $L_0^{\prime} = L_0^{\prime} \setminus \{l_{i+1}, \dots, l_{n^{\prime}-2}\}, \: n^{\prime} = i + 2 $
            \State \Call{Select features}{$L_0^{\prime}, n^{\prime}$}
        \Else
        \State \Return $B(VR_V) = \{T, l_1, \dots, l_i\}$ \Comment{the set of topological features}
        \EndIf
        \EndFor
    \EndProcedure
\State \Return \Call{Select features}{$L_0^{\prime}, \: n^{\prime}$}
\end{algorithmic}
\end{algorithm}

Let us briefly go through the major ideas of the algorithm. As we fix the edge cases, we are interested in filtering a subset of $n-2$ features as irrelevant noise. Thus, on each internal iteration, the entropy of the barcode obtained by replacing the first $i$ bars of $L_i^{\prime}$ by $i$ bars that maximize the entropy of the empirical distribution is evaluated.
Then, the idea is to iteratively apply the procedure until the estimated bar length $l_i^{\prime}$ leads to a decrease in its length, thus increasing the probability of the longest bar $T$. However, if a neutralization of the bar $l_i$ leads to an increase in its neutralized estimate $l_i^{\prime}$, then $p_n^{(i-1)} > p_n^{(i)}$, and the features from $l_{i+1}$ to $l_{n^{\prime}-2}$ are regarded as noise. At the same time, $Q$ measures the maximum number of topological features for the observed distribution and serves as a boundary for the selection process. For a much more rigorous derivation see \cite{Atienza}.

As the selected features are generally defined as the empirical distribution of distances between selected embeddings obtained after a hidden layer, we can use it to incorporate persistent entropy of this distribution as the regularization term for the loss function. 

Thus, our regularization loss is defined as follows:
\begin{center}
    $\mathcal{L}_{ent} = -\sum_{i=1}^{N \mathbb{1}[i \in L^{\prime}]} \frac{l_i}{S_{L^{\prime}}} \log(\frac{l_i}{S_{L^{\prime}}}), \: L^{\prime} = \{T, \dots, l_i\}$
\end{center}

As the term is bounded, the overall minimization problem remains well-defined.

\section{Experiments}

In order to evaluate our reasoning we test two hypotheses:
\begin{enumerate}
    \item An addition of the regularization term improves the isotropy of the latent space.
    \item The feature selection step is essential to the effectiveness of the approach, as it exploits the underlying structure encoded in representations.
\end{enumerate}

We test them by fine-tuning bert-base-uncased\footnote{https://huggingface.co/google-bert/bert-base-uncased} \cite{Devlin} and roberta-base\footnote{https://huggingface.co/FacebookAI/roberta-base} \cite{Liu} on the MRPC and CoLA datasets from the GLUE benchmark \cite{Wang} with the batch size set to 64. We collect the contextual embeddings for the regularization loss into a matrix of size $N \times D$ by taking the [CLS] token from the last hidden layer. Anisotropy values are tracked along the training process. We also track the anisotropy of the centered embeddings, which corresponds to the eigenvalues of the covariance matrix.

We use the Adam optimizer \cite{Kingma} with linear warmup for $10\%$ of training steps. The exact parameters used for fine-tuning are shown in Table \ref{tab1}. We repeat each experiment 5 times and report the average together with the standard deviation of calculated metrics over the runs.

\begin{table}
\centering
\caption{Fine-tuning parameters for classification tasks}\label{tab1}
\begin{tabular}{|l|r|r|r|}
\hline
Task & Learning rate & Weight decay & Epochs\\
\hline
BERT-MRPC & 8e-5 & 8e-6 & 10\\
RoBERTa-MRPC & 8e-5 & 8e-6 & 10\\
BERT-COLA & 5e-5 & 5e-6 & 10\\
RoBERTa-COLA & 5e-5 & 5e-6 & 10\\
\hline
\end{tabular}
\end{table}

As the end task is classification, instead of applying $\mathcal{L}_{ent}$ to the whole space, we split embeddings based on classification label, and apply the loss to them separately, thus aiming to improve isotropy group-wise. Applying it to the whole space seems more appropriate during pretraining, but not in a classification task, as it will push distant clusters towards each other instead of keeping them distinct. As we want to increase the entropy of the obtained distribution, we take it with a minus sign.

The resulting objective is defined as follows: 
\begin{center}
    $\mathcal{L} = \mathcal{L_{CE}} - \sum_{i=1}^{N}\mathcal{L}_{ent}\mathbb{1}_{[\text{class=i}]}$
\end{center}

\section{Results}

\begin{figure}%
    \centering
\adjustbox{trim=0.5cm 0.4cm 0.5cm 1.2cm}{%
\subfloat
{{\includegraphics[width=0.5\textwidth]{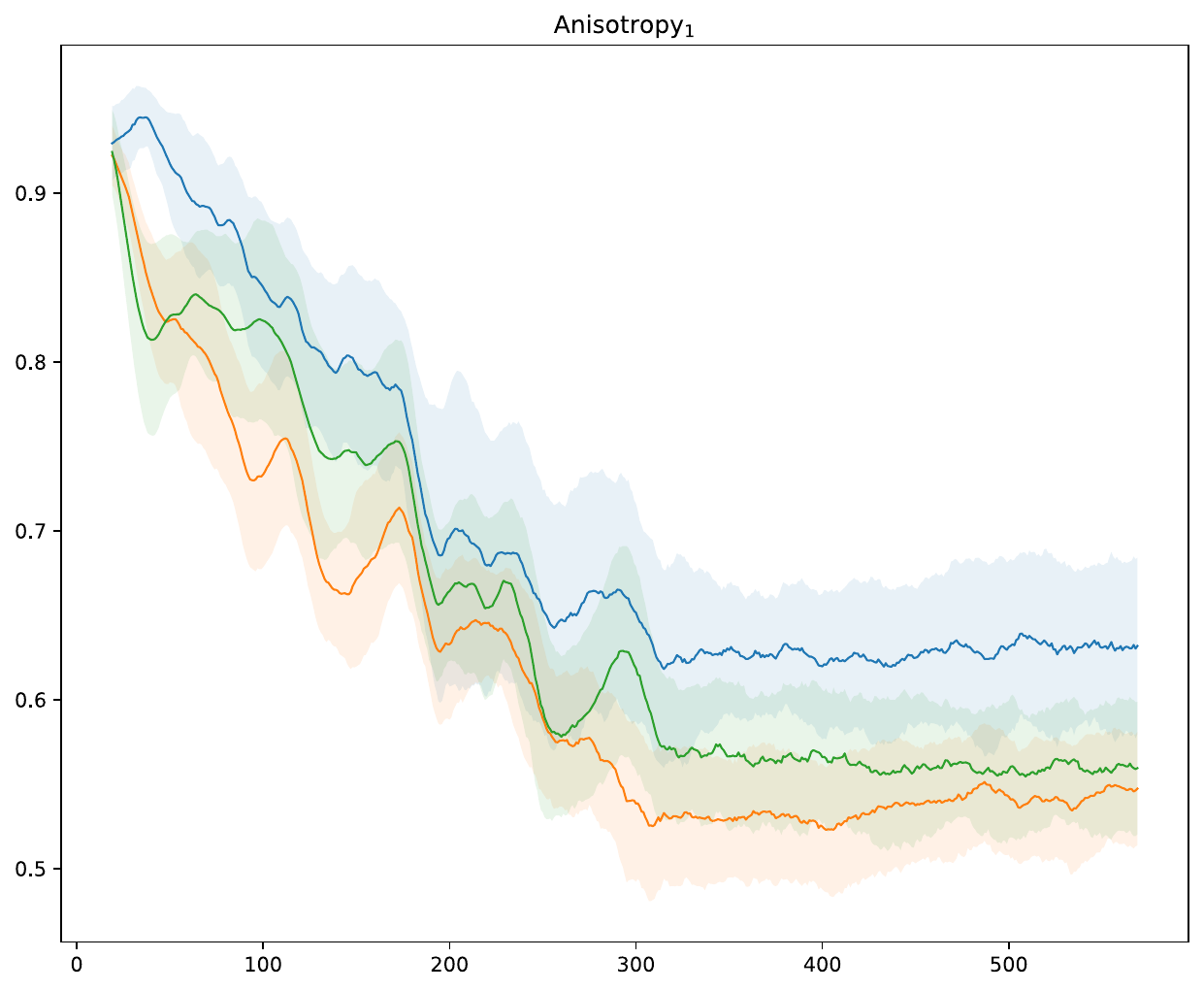} }}%
}
    \quad
\adjustbox{trim=0.5cm 0.4cm 0.5cm 1.2cm}{%
    \subfloat
    {{\includegraphics[width=0.5\textwidth]{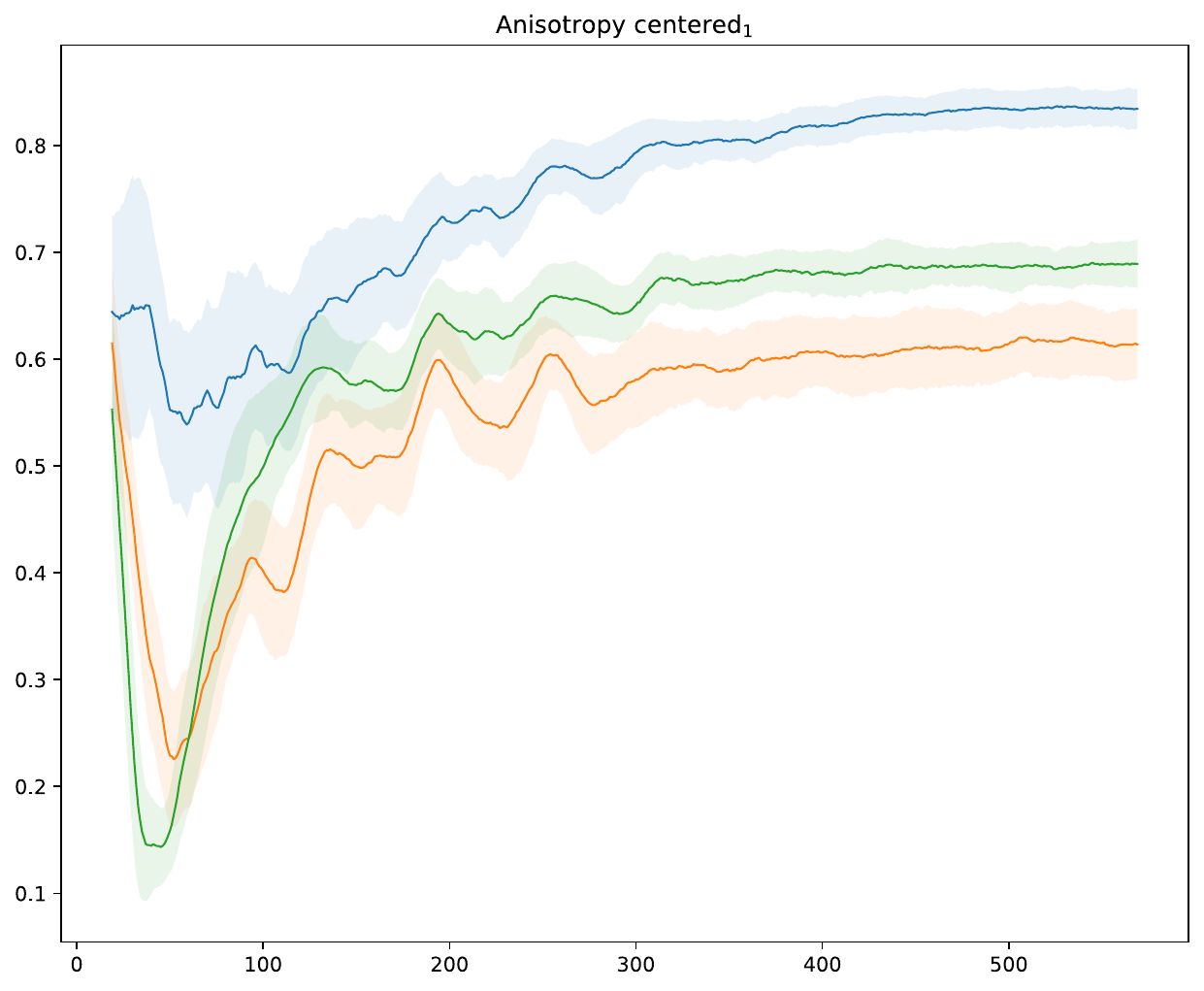} }}%
}
\\
\adjustbox{trim=0.5cm 0.4cm 0.5cm 0.5cm}{%
\subfloat
{{\includegraphics[width=0.5\textwidth]{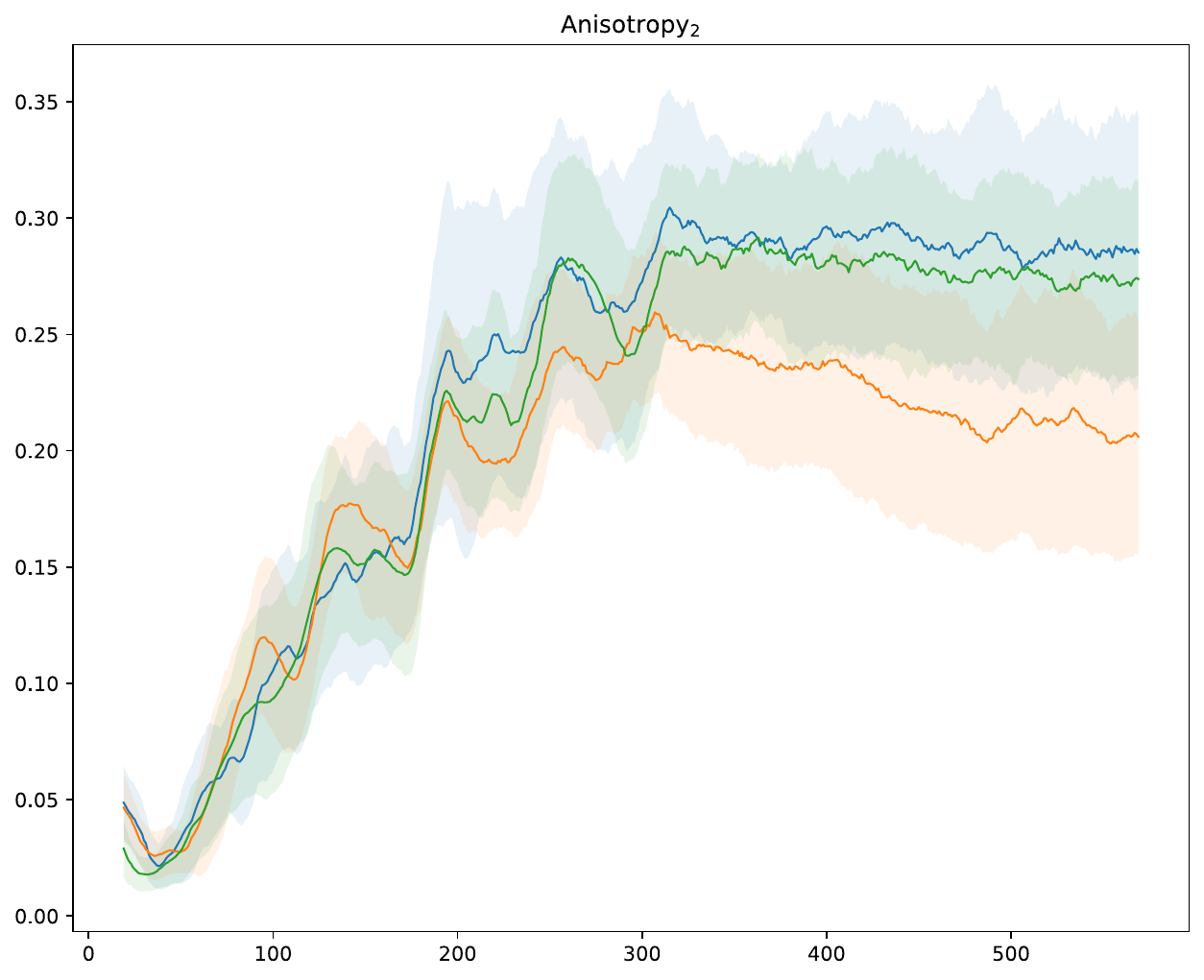} }}%
}
    \quad
\adjustbox{trim=0.5cm 0.4cm 0.5cm 0.5cm}{%
    \subfloat
    {{\includegraphics[width=0.5\textwidth]{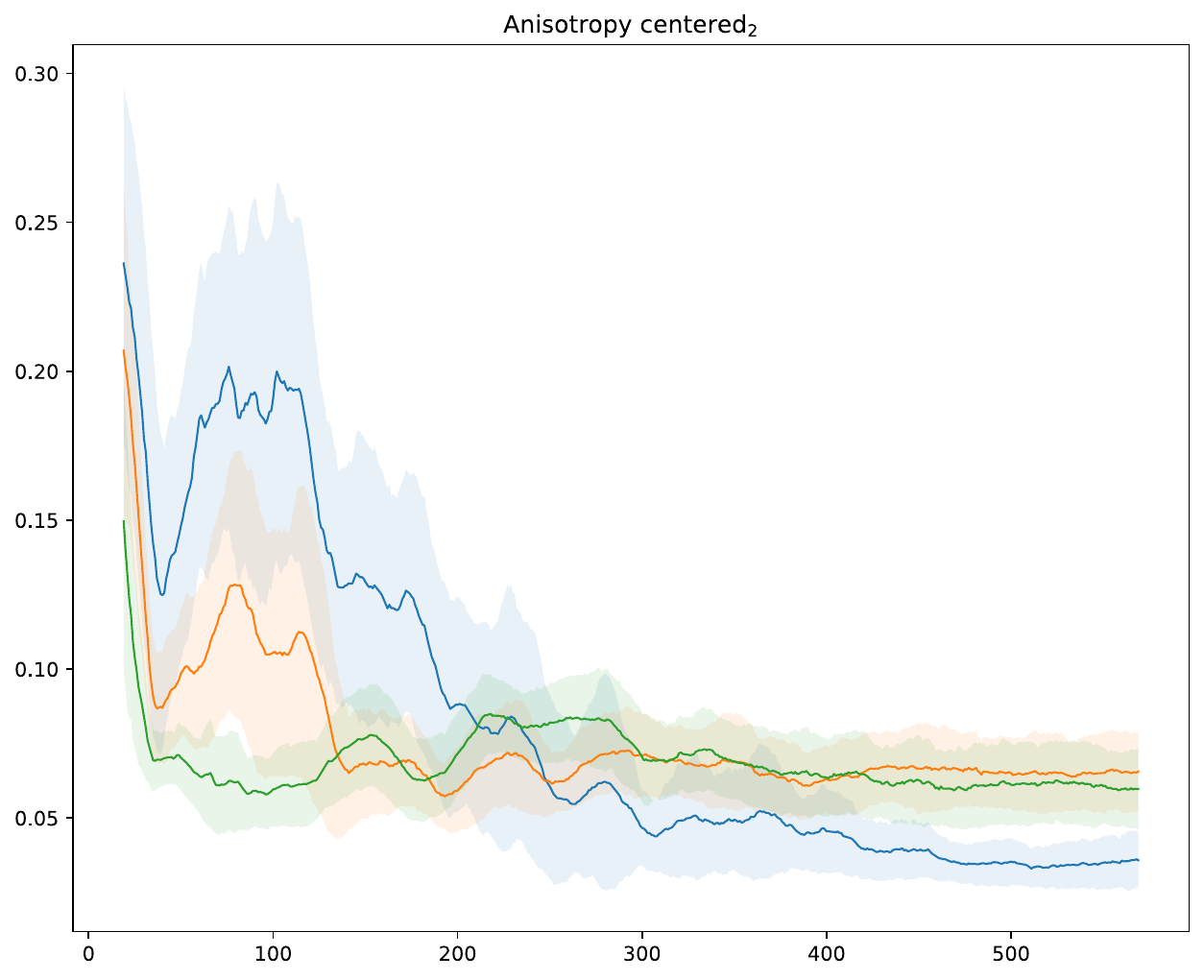} }}%
}
\\
\adjustbox{trim=0.5cm 0.4cm 0.5cm 0.5cm}{%
\subfloat
{{\includegraphics[width=0.5\textwidth]{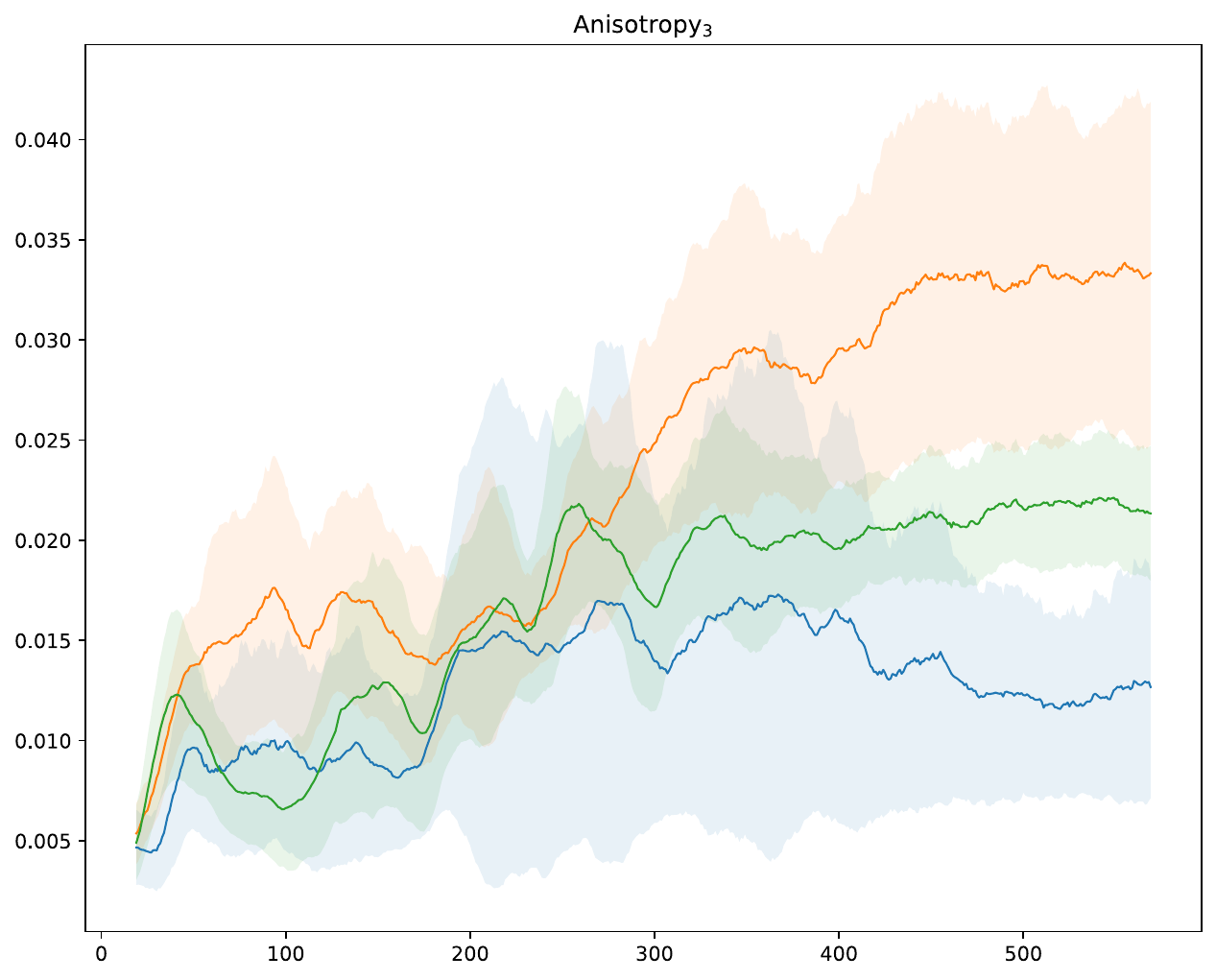} }}%
}
    \quad
\adjustbox{trim=0.5cm 0.4cm 0.5cm 0.5cm}{%
    \subfloat
    {{\includegraphics[width=0.5\textwidth]{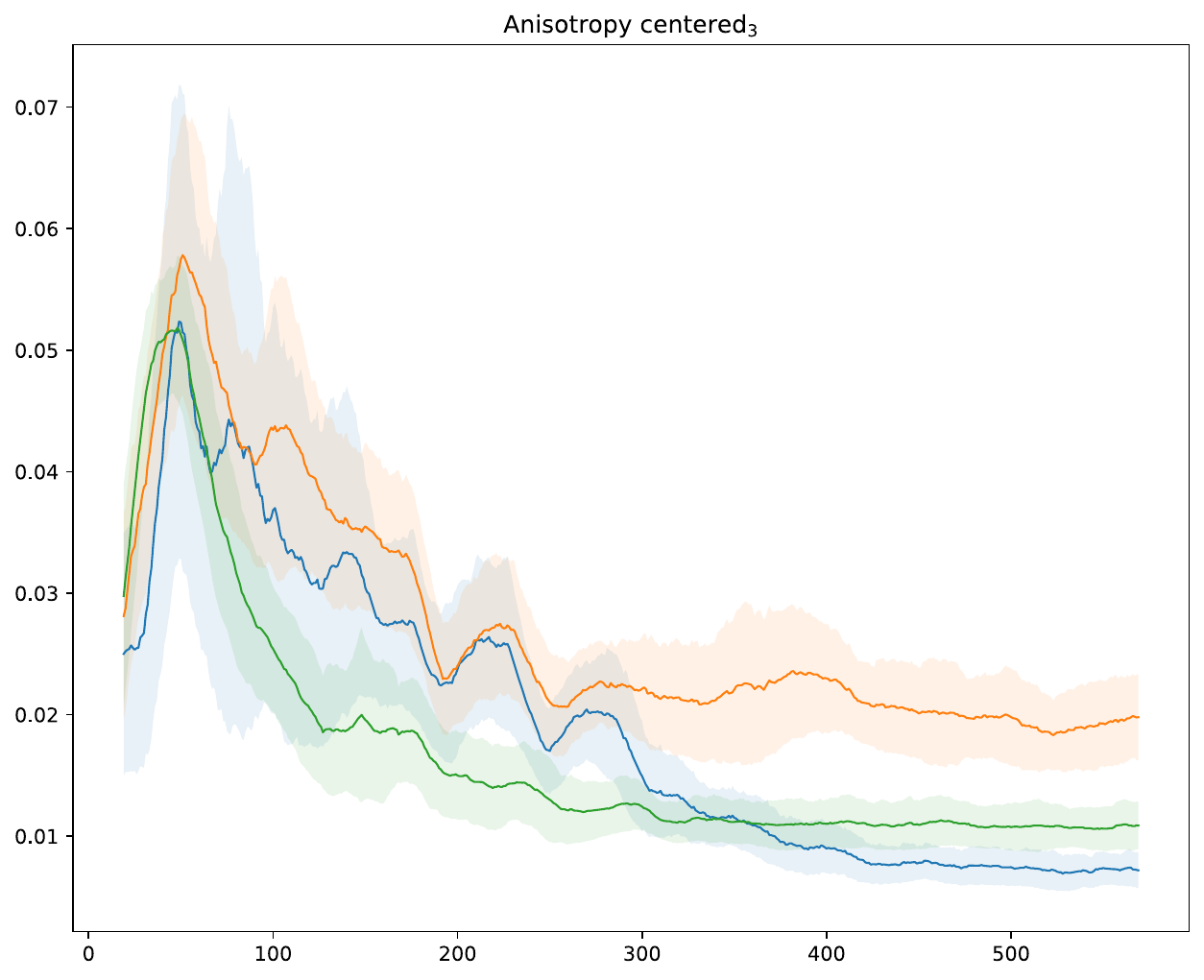} }}%
}

    \vspace{10pt}
{\color{graph_cl_1} \rule{0.5cm}{1mm}} No reg.
{\color{graph_cl_2} \rule{0.5cm}{1mm}} Ent. loss (selected barcodes)   
{\color{graph_cl_3} \rule{0.5cm}{1mm}} Ent. loss (all barcodes)

\caption{Anisotropy profiles of RoBERTa models during fine-tuning on MRPC}%
\label{fig:1}%
    
\end{figure}

We observe that the anisotropy of the latent space without regularization not only persists, but rises during fine-tuning (see Fig. \ref{fig:1}). This can be partly attributed to the emergence of a certain direction, which is aligned with the normal of the hyperplane separating the classes.
However, our regularization loss reduces the extent of anisotropy and serves as an obstacle that prevents embeddings from collapsing onto a few directions.

\begin{figure}%
    \centering
\adjustbox{trim=0.5cm 0.5cm 0.5cm 0.5cm}{%
    \subfloat
    {{\includegraphics[width=0.5\textwidth]{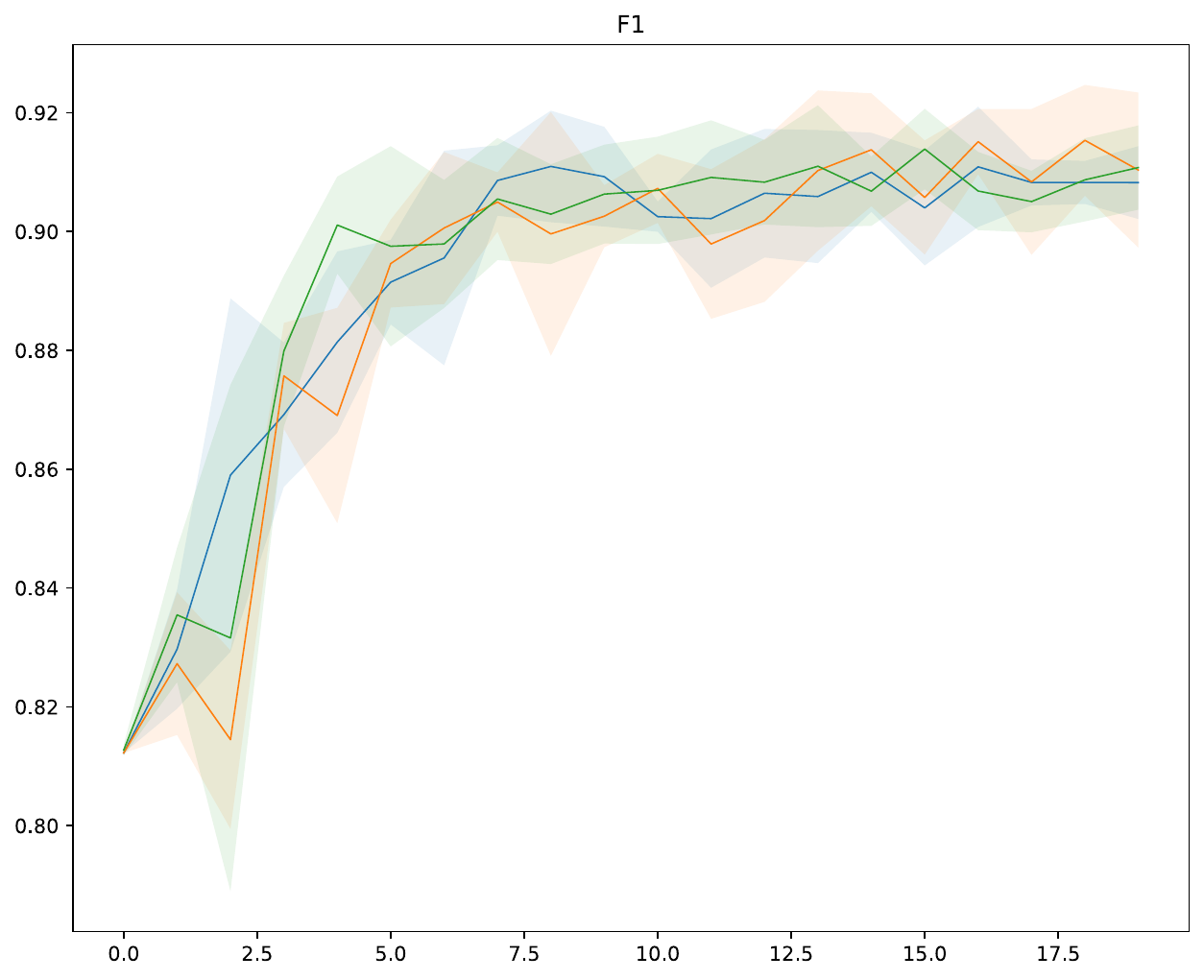} }}%
}
    \quad
\adjustbox{trim=0.5cm 0.5cm 0.5cm 0.5cm}{%
    \subfloat
    {{\includegraphics[width=0.5\textwidth]{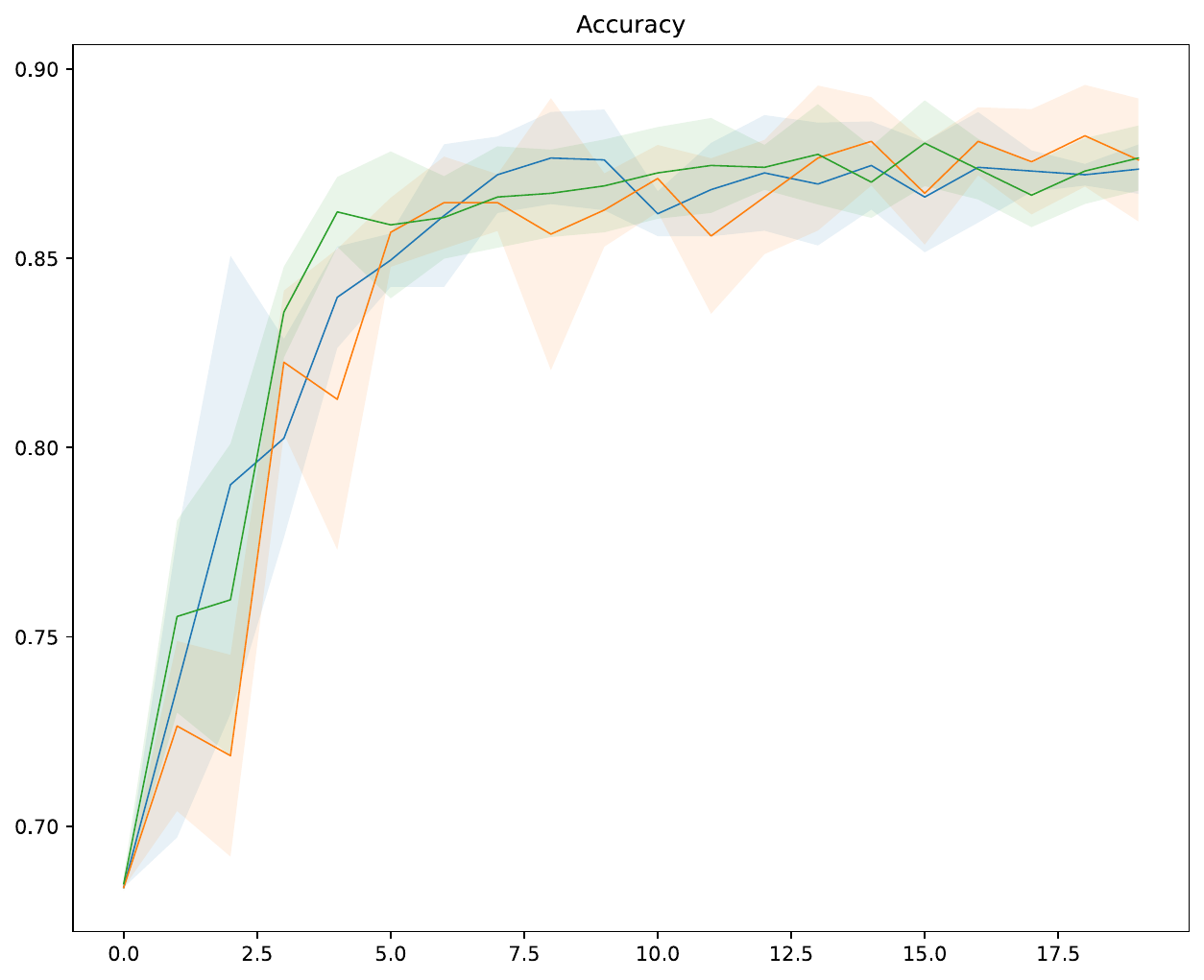} }}%
}
\\
\vspace{12pt}
    {\color{graph_cl_1} \rule{0.5cm}{1mm}} No reg. 
    {\color{graph_cl_2} \rule{0.5cm}{1mm}} Ent. loss (selected barcodes)   
    {\color{graph_cl_3} \rule{0.5cm}{1mm}} Ent. loss (all barcodes) 
    
    \caption{MRPC validation metrics of  RoBERTa models during fine-tuning}%
    \label{fig:2}%
\end{figure}

This not only leads to isotropy improvements (see Table \ref{tab3}) but also to slight growth in generalization (see Fig. \ref{fig:2}, Table \ref{tab2}). The values reported in the tables are obtained by averaging over the last 30\% of training, when the models have already converged, and over 5 independent runs.


We also observe that feature selection indeed has a substantial effect on both isotropy and generalization. As we don't discriminate between inter- and intra-cluster points if we skip the selection step, the lengths of the resulting spanning tree edges obtained at the entropy maximization step appear to be smaller. This leads to higher anisotropy as we fail to mitigate the contribution of the most prominent directions. The destruction of the cluster-wise structure leads to poorer generalization compared with selected barcodes. 

\begin{table}
\centering
\caption{Classification metrics for the experiments}\label{tab2}
\begin{tabular}{|l|wr{2.5cm}|wr{3cm}|wr{3cm}|}
\hline
Task & No reg. & Ent. loss (sel. bars) & Ent. loss (all bars)\\
\hline
BERT-MRPC & $\mathbf{0.892 \pm 0.002}$ & $0.889 \pm 0.001$ & $0.892 \pm 0.002$ \\
RoBERTa-MRPC & $0.908 \pm 0.002$ & $\mathbf{0.911 \pm 0.003}$ & $0.908 \pm 0.002$ \\
BERT-COLA & $0.575 \pm 0.002$ & $\mathbf{0.588 \pm 0.003}$ & $0.572 \pm 0.002$ \\
RoBERTa-COLA & $0.608 \pm 0.003$ & $\mathbf{0.609 \pm 0.002}$ & $0.609 \pm 0.004$ \\
\hline
\end{tabular}
\end{table}

It is nonetheless interesting that such destruction doesn't lead to a gradual decrease in downstream performance over the course of training, and the models perform mostly on par with the ones trained without regularization. 

\begin{table}[h]
\centering
\caption{Anisotropy values for the experiments}\label{tab3}
\begin{tabular}{|l|l|r|r|r|}
\hline
Task & Ani. component & No reg. & Ent. loss (sel. bars) & Ent. loss (all bars)\\
\hline
\multirow{6}{*}{BERT-MRPC} & $\text{Anisotropy}_1$ & 0.679 & 0.485 & 0.541 \\
& $\text{Anisotropy}_2$ & 0.125 & 0.098 & 0.165 \\
& $\text{Anisotropy}_3$ & 0.023 & 0.042 & 0.028 \\
& $\text{Ani. centered}_1$ & 0.733 & 0.483 & 0.576 \\
& $\text{Ani. centered}_2$ & 0.034 & 0.057 & 0.050 \\
& $\text{Ani. centered}_3$ & 0.017 & 0.024 & 0.017 \\
\hline
\multirow{6}{*}{RoBERTa-MRPC} & $\text{Anisotropy}_1$ & 0.629 & 0.540 & 0.559 \\
& $\text{Anisotropy}_2$ & 0.287 & 0.214 & 0.276 \\
& $\text{Anisotropy}_3$ & 0.012 & 0.032 & 0.021 \\
& $\text{Ani. centered}_1$ & 0.832 & 0.611 & 0.686 \\
& $\text{Ani. centered}_2$ & 0.036 & 0.065 & 0.061 \\
& $\text{Ani. centered}_3$ & 0.007 & 0.019 & 0.010 \\
\hline
\multirow{6}{*}{BERT-COLA} & $\text{Anisotropy}_1$ & 0.647 & 0.499 & 0.502 \\
& $\text{Anisotropy}_2$ & 0.099 & 0.076 & 0.078 \\
& $\text{Anisotropy}_3$ & 0.028 & 0.028 & 0.023 \\
& $\text{Ani. centered}_1$ & 0.660 & 0.468 & 0.478 \\
& $\text{Ani. centered}_2$ & 0.038 & 0.037 & 0.032 \\
& $\text{Ani. centered}_3$ & 0.018 & 0.022 & 0.020 \\
\hline
\multirow{6}{*}{RoBERTa-COLA} & $\text{Anisotropy}_1$ & 0.581 & 0.534 & 0.542 \\
& $\text{Anisotropy}_2$ & 0.277 & 0.222 & 0.221 \\
& $\text{Anisotropy}_3$ & 0.016 & 0.021 & 0.018 \\
& $\text{Ani. centered}_1$ & 0.700 & 0.512 & 0.510 \\
& $\text{Ani. centered}_2$ & 0.056 & 0.073 & 0.063 \\
& $\text{Ani. centered}_3$ & 0.013 & 0.019 & 0.017 \\
\hline
\end{tabular}
\end{table}

Overall, the addition of entropy regularization paired with topological feature selection leads to improvements both in terms of isotropy and generalization without the need for reparametrizing the model, and is suitable for fine-tuning scenarios.

\section{Conclusion}

In our work we presented a novel regularization loss aimed at improving the isotropy of the latent space, based on simplicial geometry and topological data analysis. We empirically evaluated its effect in a fine-tuning scenario and found that it leads to an improvement in preserving the higher isotropy of the latent space, while improving the downstream metrics, which was reported to be inconsistent for methods that rely on reparametrization of the space instead of exploiting existing structures \cite{Ding}. On the contrary, our approach performs well even in the absence of a large training dataset and doesn't rely on retraining the model. It also doesn't add any inference overhead. It should be noted that although we apply our regularization loss to the language model's latent space, the general construction is actually model-agnostic and applicable to any optimization problem where the data has a point cloud structure. We leave an exploration of its usefulness on other model architectures and tasks for future work.

\end{document}